%% file: main.tex
\documentclass[graybox,envcountchap]{SNmult}

\input{preamble}

\makeindex

\begin{document}

\title*{Artificial Intelligence for Modeling \& Simulation in Digital Twins}

\author{Philipp Zech and Istvan David}

\institute{Philipp Zech \at University of Innsbruck, Austria \email{philipp.zech@uibk.ac.at}
\and Istvan David \at McSCert, McMaster University, Canada \email{istvan.david@mcmaster.ca}}

\maketitle

\abstract{The convergence of \ms{} (\MS{}) and artificial intelligence (AI) is leaving its marks on advanced digital technology. Pertinent examples are digital twins (DTs)---high-fidelity, live representations of physical assets, and frequent enablers of corporate digital maturation and transformation. Often seen as technological platforms that integrate an array of services, DTs have the potential to bring AI-enabled \MS{} closer to end-users. It is, therefore, paramount to understand the role of \MS{} in DTs, and the role of digital twins in enabling the convergence of AI and \MS{}. To this end, this chapter provides a comprehensive exploration of the complementary relationship between these three. We begin by establishing a foundational understanding of DTs by detailing their key components, architectural layers, and their various roles across business, development, and operations. We then examine the central role of \MS{} in DTs and provide an overview of key modeling techniques from physics-based and discrete-event simulation to hybrid approaches. Subsequently, we investigate the bidirectional role of AI: first, how AI enhances DTs through advanced analytics, predictive capabilities, and autonomous decision-making, and second, how DTs serve as valuable platforms for training, validating, and deploying AI models. The chapter concludes by identifying key challenges and future research directions for creating more integrated and intelligent systems.}

\section{Introduction}

\index{digital twin|(}Digital twins (DT) are high-fidelity virtual representations of physical assets~\cite{rasheed2020digital}. They allow for advanced engineering scenarios by implementing two key mechanisms: computational reflection, through which they reflect the behavior exhibited by the physical asset; and closed-loop control, through which they can modify and precisely manipulate the physical asset. (Sometimes, DTs lack control capabilities, reducing them to mere digital shadows or runtime models. Here, we consider a DT inclusive of a control loop.) In the typical DT terminology, the physical asset is referred to as the \textit{physical twin}~\cite{kritzinger2018digital} or \textit{actual twin}~\cite{tao2019digital}---underscoring that there are two twins in this setup (digital and physical) and changes in any of the twins will be eventually reflected in the other.
This bi-directional coupling between twins is particularly useful in situations when computer-aided reasoning is required in support of managing the physical twin, such as automated optimization, real-time reconfiguration, and intelligent adaptation of the physical twin.\index{digital twin|)}
Such scenarios are enabled by \ms{} (\MS{}).

Models describe how the real system is structured and how it behaves. Through the mechanism of abstraction, models capture the properties of the system that are of interest. For example, a physics model may contain information about the temperature of a machine, while a Petri net may focus on the discrete states a machine exhibits. Through this simplified view on the enormously complex real world, models can be used in simulators---programs that are able to enact various scenarios of interest and by that, predict the future states of the system.
DTs use \MS{} to derive the control strategy that keeps the physical twin within a desired operational state---e.g., the most energy-efficient one or most performant one---under changing conditions.

Of course, physical assets subject to digital twinning exhibit high complexity and their detailed modeling may be infeasible for a human. To understand the behavior of physical systems, one needs to experiment with them---e.g., measure their physical properties, or observe them under different environmental conditions---to collect evidence from which models can be constructed~\cite{cederbladh2025reasonable}. This experimentation may be a time-consuming endeavor as systems complexity increases and one wishes to isolate causal relationships (``\textit{if I change this, that other thing changes}'') which are essential in building models.
To circumvent this issue, machine learning and artificial intelligence (AI) are increasingly more often called upon as modeling aids, as discussed in Chapter 1.
AI models can be trained by processing voluminous data and crunching the numbers. The resulting AI model, in turn, will encode the statistical behavior of the twinned system, which can be subsequently used more or less just like any ordinary model in simulation scenarios.

This chapter explores how DTs can leverage \MS{} and AI and provide a platform for advancing and deepening their integration. To this end, \secref{sec:dt} continues with outlining the foundational components and architectural aspects of DTs. \secref{sec:ms} dives into the role of \MS{} in DTs and provides an overview the main \MS{} approaches used in DTs. \secref{sec:ai} then discusses the synergistic nature of DTs and AI, i.e., how AI enhances DTs, and how DTs accelerate AI development. \secref{sec:outlooks} concludes this chapter by summarizing the main challenges and recommending future directions by providing a roadmap for future research and development.

\index{digital twin|(}
\section{Foundations of Digital Twins}\label{sec:dt}

DTs have emerged as a disruptive paradigm in the industrial landscape, unlocking novel capabilities for monitoring, analyzing, and optimizing physical systems~\cite{zech2025empirical}. The concept of a DT is rooted in NASA's Apollo programs, tracing back to the 1960s. DTs were intended to respond to spacecraft failures, dubbed as ``living models'', i.e., simulators that evaluate the health of the spacecraft based on continuously ingested data. A few decades later, similar ideas emerged in the product lifecycle management (PLM) field, labeled as the ``conceptual ideal for PLM'' and later, as ``the information mirroring model''~\cite{grieves2017digital}.
The concept is well-illustrated by a DT of an aircraft engine that is used for health management~\cite{wu2021framework}. The physical asset is an in-service jet engine monitored by Internet of Things (IoT) sensors. The virtual counterpart is a predictive model (cf.~a Long-Short-Term-Memory (LSTM) model), that is dynamically updated with live operational data. This continuous synchronization provides an accurate, up-to-date estimation of the engine’s health and its Remaining Useful Life (RUL)~\cite{wu2021framework}.

The first definition of a DT appears in a 2010 NASA roadmap document~\cite{shafto2010draft}: ``an integrated multiphysics, multiscale, probabilistic simulation of an as-built vehicle or system that uses the best available physical models, sensor updates, fleet history, etc., to mirror the life of its corresponding flying twin.'' The DT is also considered ``ultra-realistic,'' thanks to high-fidelity physical models, real-time sensor data, and the ability to aggregate maintenance history and fleet data.

Today, DTs have vastly outgrown the initial ideas of telemetry and PLM, and are drivers of complex socio-technical systems. Some of the key applications of DTs include automated optimization, real-time reconfiguration, and intelligent adaptation of physical systems. These capabilities and the simplicity of the DT concept render DTs particularly appealing in settings where cyber-physical systems have to be controlled in a computer-automated fashion, such as uncrewed automotive systems~\cite{ramdhan2025engineering}, smart ecosystems~\cite{michael2024digital}, and smart grids~\cite{liu2026introduction}.

In 2017, Gartner reported that DTs entered the innovation trigger phase of their annual hype cycle report, i.e., a rapid growth of interest in the concept and subsequent adoption is imminent in the next 5--10 years.\footnote{\url{https://www.gartner.com/en/documents/3768572}} Due to an unusually rapid progress, the 2018 edition of the same report already placed DTs at the peak level of (inflated) expectations.\footnote{\url{https://www.gartner.com/en/documents/3885468}} Later reports do no longer mention DTs, and it is widely accepted that DTs have reached the steady level of maturity.
Recent market evaluations report a compound annual growth rate of over 60\% in the global digital transformation market, of which digital twins are key drivers. This rapid expansion illustrates the technology's ability to provide significant operational enhancements: General Electric (GE) indicates that digital technologies have facilitated reductions of 10-20\% in maintenance expenses, 20\% in unanticipated downtime, and 15\% enhancements in energy efficiency within their industrial applications~\cite{gedigital2023digital}. Such success stories are not uncommon as DTs have transformed complete industries and sectors by now.

\subsection{Key Components of Digital Twins}
\label{ssec:dts}

\begin{figure}[hbt]
    \centering
    \includegraphics[width=.55\textwidth]{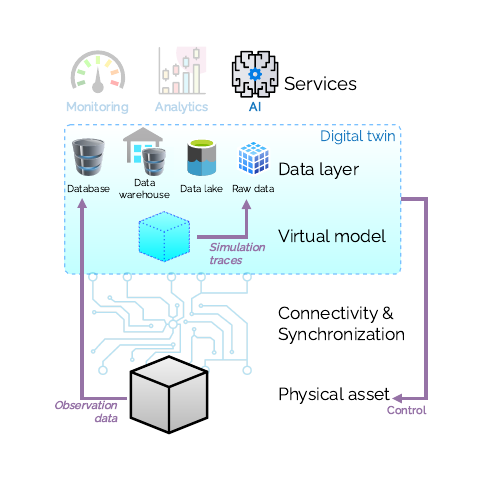}
    \caption{The key components of a DT include the \emph{physical asset} and \emph{virtual model}, which exchange data via the \emph{connectivity~\&~synchronization} layer. On top of this, the \emph{data layer} manages data collected from both the physical asset and from simulations of the virtual model.}
    \label{fig:dt_comp}
\end{figure}

As shown in \figref{fig:dt_comp}, a DT system integrates four key components: (i) the physical asset on the physical side; and on the digital side, (ii) a set of digital models, (iii) a data management layer, and (iv) robust connectivity and synchronization middleware to connect physical and digital components.
In the following, we discuss these components.

\index{digital twin!physical asset|(}
\subsubsection{Physical asset: The real-world entity}

The physical asset is the tangible, real-world entity that is subject to digital twinning~\cite{grieves2017digital}. 
This component may encompass a spectrum of scales and complexities, ranging from 
elementary machine components (e.g., turbine blades, electronic control units, hydraulic actuators) 
to sophisticated integrated subsystems (e.g., complete aircraft propulsion systems, automated manufacturing 
lines, intelligent building infrastructure)~\cite{tao2018digital}.
The physical asset functions as the authoritative reference point for the DT ecosystem, providing both the structural template for modeling the physical asset's manifestation and the continuous source of operational telemetry~\cite{lu2020digital}. 

Physical systems have undergone significant transformation through the integration of  extensive instrumentation, featuring diverse sensor arrays that capture multidimensional operational parameters with high fidelity~\cite{boschert2016digital}. Typical operational parameters include the following.

\begin{description}
    \item[\textbf{Operational conditions}] Such as, temperature gradients, pressure differentials, humidity fluctuations, fluid levels, and chemical compositions.
    \item[\textbf{Performance metrics}] Such as, production throughput, energy efficiency coefficients, quality conformance indicators, cycle time measurements.
    \item[\textbf{Environmental factors}] Such as, vibrational signatures, acoustic emissions, electromagnetic interference, particulate concentrations.
    \item[\textbf{Maintenance indicators}] Such as, component wear patterns, material fatigue progression, structural integrity deviations, anomalous behavioral signatures.
\end{description}

The integration of IoT technologies~\cite{david2025modeling} has further improved the capabilities of physical assets to generate rich, high-dimensionality data streams~\cite{qi2021enabling}.
According to \citet{gartner2023forecast}, the global deployment of connected IoT devices has reached approximately 43 billion units in 2025---a threefold increase from 2018 levels following an exponential growth trajectory. This reflects the accelerating trend toward comprehensively instrumented and networked physical assets across industrial sectors.

The sophistication of asset instrumentation correlates with the fidelity and utility of the DT, as it establishes a deterministic relationship between sensor deployment strategy and achievable DT capabilities~\cite{kritzinger2018digital}. Today, physical assets often provide self-diagnostic capabilities through smart sensors that perform edge analytics, which enables data processing and anomaly detection at the source---a capability that significantly enhances the responsiveness and efficiency of the DT system~\cite{rasheed2020digital}.

\phantom{}

For example, Zech et al.~\cite{zech:annsim:2024} describe a test rig of a two-story building as the physical asset, which is instrumented with a pervasive network of wired and wall-embedded RFID sensors. These devices generate continuous, high-fidelity data streams that capture key operational conditions, such as air temperature, humidity, and wall moisture levels. This application is a typical example of how a physical asset's extensive sensor instrumentation provides the fundamental operational telemetry required to create and sustain a high-fidelity DT.\index{digital twin!physical asset|)}

\index{digital twin!virtual model|(}
\subsubsection{Virtual model: The simulation-based representation}

The virtual model constitutes the digital representation of the physical asset, capturing its physical 
characteristics, behaviors, and operational parameters. This model serves as the computational environment 
where simulations, analyses, and optimizations can be performed. The virtual model typically comprises multiple 
integrated elements~\cite{tao2022digital}, including geometric representation (e.g., 3D CAD models), physical behavior models (e.g., finite element models, computational fluid dynamics), functional models (e.g., system dynamics, control logic), process models (e.g., operational workflows, manufacturing sequences). An illustrative example of integrating various model types is the DT of a port structure by Jayasinghe et al.~\cite{jayasinghe2024innovative}. The virtual model employs a hybrid approach: a high-fidelity, physics-based Finite Element Model (FEM) is first used to simulate the port structure's behavior under various loads, thereby generating data which then is used to to train an Artificial Neural Network (ANN). The trained ANN then serves as a computationally efficient surrogate model, which provides real-time predictions of the entire port structure's behavior from sensor data. With the FEM model alone, real-time prediction would be infeasible~\cite{jayasinghe2024innovative}.
This example clearly illustrates that the fidelity of models varies based on the application requirements and computational constraints. 
High-fidelity models may incorporate detailed physics-based simulations that accurately represent the asset's 
behavior under various conditions, while reduced-order models might prioritize computational efficiency for 
real-time applications~\cite{boschert2016digital}.

The virtual model must be sufficiently comprehensive to represent the physical asset across its entire life cycle, from design and manufacturing to operation and maintenance, enabling what they term \emph{product lifecycle DT}~\cite{tao2018digital}. In addition to supporting the usual predictive maintenance tasks of a DT, a product lifecycle DT also facilitates optimization strategies and decision-making processes across the asset’s entire lifespan. Moreover, the incorporation of data from real-time sensors into the virtual model ensures continuous updates and refinements, increasing the model’s accuracy and relevance over time~\cite{kritzinger2018digital,tao2018digital}. In addition, integrating both physical and functional data streams into the virtual model is paramount~\cite{lu2020digital}. These streams allow the virtual model to reflect real-world performance, moving beyond simple design representations to dynamic, operational models that evolve and adapt in response to changing conditions.\index{digital twin!virtual model|)}

\index{digital twin!data layer|(}
\subsubsection{Data layer: Analytics, machine learning, and automation}

The data layer enables transformation of raw data from the physical asset into actionable insights, predictions, and autonomous decisions, enhancing the functionality and intelligence of the DT. This layer plays a pivotal role in enabling the DT to not only represent the physical asset’s current state but also predict future states, optimize operations, and recommend or autonomously implement optimal actions. The data layer establishes the bridge between the data generated by the physical asset and the decisions that drive operational improvements by providing the data basis for various analytical and AI techniques, including the following.

\begin{description}
    \item[\textbf{Data acquisition and pre-processing systems}] These systems are responsible for collecting data from sensors of the physical asset. Pre-processing techniques such as filtering, normalization, and outlier removal are applied to ensure the quality and reliability of the data before further analysis~\cite{kong2021data}.
    
    \item[\textbf{Pipelines for descriptive, diagnostic, predictive, and prescriptive analytics}] Descriptive analytics focuses on understanding historical data and identifying patterns and trends~\cite{javaid2023digital}. Diagnostic analytics helps to determine the causes of problems or inefficiencies. Predictive analytics leverages historical data to forecast future asset behaviors or failures, enabling proactive maintenance and informed decision-making. Prescriptive analytics prescribes actions to identify the best course of action based on the predictions.
    
    \item[\textbf{Machine learning for pattern recognition, anomaly detection, and optimization}] Machine learning models are used to detect patterns in large datasets, identify anomalies, and predict potential failures or performance drops. These models continuously improve as more data is collected, increasing their accuracy and predictive power over time~\cite{rathore2021role}.
    
    \item[\textbf{Decision support systems and autonomous control mechanisms}] These systems use the insights generated by the analytics and AI models to support human decision-making or to act autonomously. For instance, in predictive maintenance scenarios, the system might recommend repairs or trigger automatic adjustments to the asset’s operation to avoid, thereby increasing reliability and efficiency~\cite{zhong2023overview}.
\end{description}

Through the data layer, the DT gains the capability to mirror not only the current state of the physical asset but also predict its future states and recommend or automatically implement optimal actions. Subsequent leveraging of AI and machine learning (\secref{sec:ai}) allows the DT to learn from past data, continuously improving its performance and adapting to new circumstances. This makes the DT a valuable tool for predictive maintenance, asset optimization, and real-time decision-making~\cite{bariah2024interplay}.

\phantom{}

Woo et al.~\cite{woo2025exploring} highlight the application of optimization techniques for the management of electric vehicle (EV) fleets. In their study, Woo et al.~use real-world EV charging and telemetry data from the San Francisco Bay Area to feed an optimization model. The DT's data layer implements prescriptive analytics  using mixed-integer linear programming to determine the optimal charging and discharging schedules for each vehicle. The model's overall objective is to minimize charging costs and \coo{} emissions while integrating with the power grid's dynamic pricing.
\index{digital twin!data layer|)}

\index{digital twin!connectivity|(}
\subsubsection{Connectivity~\&~synchronization: Real-time data exchange}

The connectivity~\&~synchronization layer enables bidirectional communication between the physical asset and virtual model, ensuring that virtual and physical components remain aligned. This connection is essential for DTs to accurately reflect the current state of the physical asset and enable adjustments (possibly at real-time), which can improve decision-making and operational efficiency. The effective functioning of this layer depends on the timely transmission and synchronization of data, as well as the ability to implement control actions based on insights derived from the virtual model. Key components in this layer include the following.

\begin{description}
    \item[\textbf{Communication protocols}] Communication protocols such as 5G, industrial Ethernet, and fieldbus systems like Modbus and PROFIBUS are central to enabling seamless data flow between the physical asset and its virtual counterpart~\cite{tao2018digital}. These protocols ensure the continuous communication between the DT and the physical asset, enabling real-time updates and operational decisions.
    
    \item[\textbf{Data synchronization mechanisms}] Maintaining temporal coherence between the physical and virtual systems is crucial. This layer includes synchronization mechanisms that ensure that the data from the physical asset, collected through IoT sensors and other devices, is accurately reflected in the virtual model. As \citet{kritzinger2018digital} note, accurate synchronization is essential for applications such as predictive maintenance, where real-time data helps to anticipate failures and reduce unplanned downtime. In the case of process control, the timely synchronization of data ensures operational continuity.
    
    \item[\textbf{Edge computing}] The integration of edge computing within the connectivity layer enhances the ability to process data locally, near the asset. This reduces latency and alleviates the strain on network bandwidth, which is critical when processing large volumes of real-time data. Edge computing facilitates real-time decision-making by running data analysis algorithms on-site, allowing for faster responses without depending entirely on cloud infrastructure. This capability is particularly valuable for applications requiring low-latency responses, such as manufacturing operations or autonomous systems~\cite{wang2023survey}.
    
    \item[\textbf{Security measures}] As DTs are increasingly utilized to manage critical infrastructure, ensuring the security of data exchange between the physical and virtual models is paramount. Encryption, access control, and authentication mechanisms are crucial for protecting data integrity and preventing unauthorized access, especially in industries with sensitive information, such as energy and healthcare. The implementation of robust security protocols guarantees that the real-time data exchanged between physical and digital systems remains secure.
\end{description}

The ability to exchange data in real time is indispensable for applications requiring immediate responses. In predictive maintenance, for example, continuous monitoring of physical assets is crucial to detect early signs of wear or failure, preventing costly downtime. Real-time synchronization allows the DT to trigger maintenance alerts based on up-to-date data, reducing unplanned service interruptions~\cite{tao2018digital}.

In process control scenarios, the DT’s real-time data synchronization enables immediate operational adjustments. This capability leads to better resource management, such as reducing energy consumption or optimizing manufacturing workflows, based on continuous updates from the physical system~\cite{xu2023survey}.

\begin{figure}
    \centering
    \includegraphics[width=0.7\textwidth]{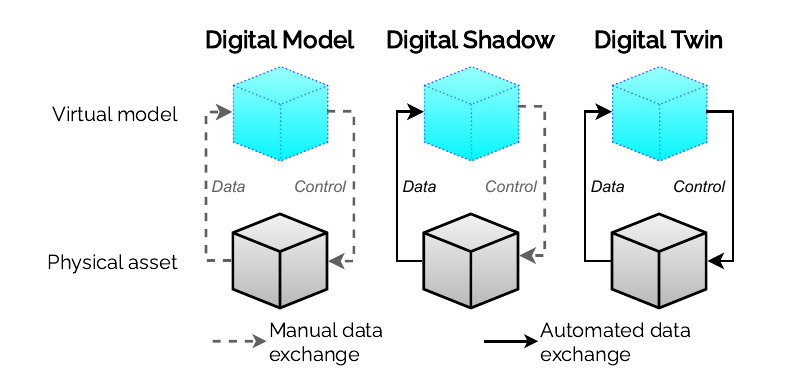}
    \caption{Digital assets classified by the level of automation in data exchange with the corresponding physical asset, in accordance with \citet{kritzinger2018digital}}
    \label{fig:dt-types}
\end{figure}

The degree of automation in the data exchange process varies widely~\cite{kritzinger2018digital}. At the lower end of the spectrum, \emph{Digital Models} rely on manual data transfer, while \emph{Digital Shadows} implement a one-way data flow where the physical asset provides data to the virtual twin. Full DTs, however, represent the highest level of automation, where the system supports bidirectional communication, allowing the virtual model to not only receive real-time data but also influence the physical asset’s behavior. This dynamic exchange enables the DT to proactively manage and optimize the asset’s performance based on real-time insights. \figref{fig:dt-types} outlines the levels of automation.

\phantom{}

A comprehensive example that integrates low-latency communication, edge computing, and security is available from \citet{zhou2021secure} in their DT for a 5G-empowered electrical distribution grid. Their DT leverages 5G communication and edge computing to process data from IoT sensors locally to enable low-latency processing and communication for real-time grid control and resource scheduling. Their approach ensures that the virtual and physical twins remain tightly synchronized. Furthermore, they integrate explicit security measures in the edge layer to identify and reject data from malicious devices to protect the data and ensure the reliability of the DT's decisions. This approach illustrates how the combination of advanced connectivity and computing technologies is key for creating secure and responsive DTs for critical infrastructures.
\index{digital twin!connectivity|)}

\subsection{The Roles of Digital Twins}

\begin{figure}
    \centering
    \includegraphics[width=0.4\linewidth]{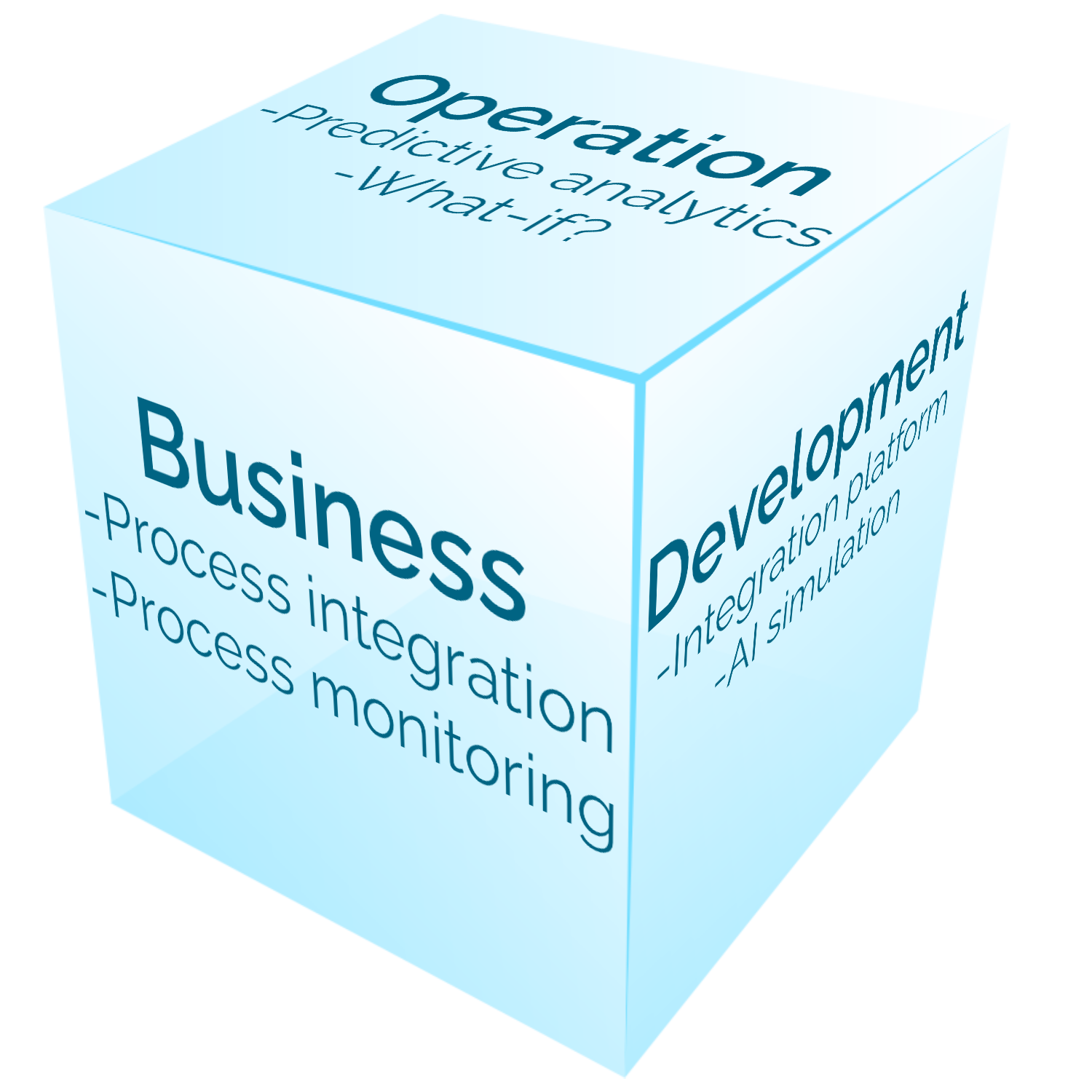}
    \caption{Many views of the same concept: DTs play various operational, business, and development roles in companies, providing a frequent source of misunderstandings due to different interpretations}
    \label{fig:cube}
\end{figure}

DTs play multiple roles across the operational, developmental, and business layers of their stakeholders (\figref{fig:cube}). DTs typically play different roles for different stakeholder groups, which is a source of frequent misunderstandings. Some see a DT as development aid, e.g., to provide a platform for technical integration, while others see it as a business aid, e.g., to integrate with business processes and monitor them.

\subsubsection{Operation}

\begin{description} 
    \item[\textbf{Operational aid}] The traditional purpose, supporting predictive analytics, condition monitoring, and operational optimization.
    \item[\textbf{Virtual experimentation and simulation sandbox}] Enables risk-free testing, evaluation, optimization of operations, and investigating \emph{what-if} scenarios.
\end{description}

\subsubsection{Development}

\begin{description}
    \item[\textbf{Technical integration platform}] Serves as a modular foundation for integrating \MS{}, AI, and heterogeneous data sources.
    \item[\textbf{AI simulation}] Provides means for generating high-quality data using simulation for training AI models.
\end{description}

\subsubsection{Business}

\begin{description}
    \item[\textbf{Decision-support system}] Facilitates integration with business processes (e.g., RAMI 4.0~\cite{rami-website}).
    \item[\textbf{End-to-end process monitoring}] Enables comprehensive digital threads and value chain transparency.
\end{description}

The examples in the previous sections illustrate these distinct but often overlapping roles. The aircraft engine DT~\cite{wu2021framework}, for example, serves an \emph{operational role} by providing predictive analytics for health monitoring and estimating RUL. The virtual model of the port structure~\cite{jayasinghe2024innovative}, which uses simulation data to train a surrogate model, highlights the \emph{development role} by acting as a platform for model development. Finally, the vehicle-to-grid optimization system~\cite{woo2025exploring} illustrates the \emph{business role} with the DT acting as a decision-support system that predicts optimal actions to achieve economic objectives.

\subsection{The Synergy of \MS{} and AI in DTs}

The integration of \MS{} and AI within DTs marks substantial improvement to managing and optimizing complex systems. The synergy between \MS{} and AI leverages the strengths of each component to create a dynamic, responsive, and intelligent system capable of real-time monitoring, predictive analytics, and autonomous decision-making. According to a report by McKinsey~\&~Company~\cite{McKinsey_2024}, organizations combining DT technology with 
generative AI have seen significant advancements in operational efficiency. By leveraging AI to enhance the deployment 
and optimization of DTs, companies across industries like manufacturing and energy can achieve better decision-making, 
predictive maintenance, and real-time optimization of asset performance.

\MS{} techniques provide the environment necessary to replicate and analyze the behavior of physical assets~\cite{wagg_digital_2020}. Through \MS{}, DTs can simulate various scenarios, predict outcomes, and optimize physical assets without the need for costly physical experimentation~\cite{boschert2016digital}. This is crucial for understanding system behavior and dynamics, identifying issues, and exploring optimization strategies before implementing and deploying them in the real world~\cite{lechler2019virtual}.

AI enhances the DT by introducing advanced analytics, machine learning, and autonomous control mechanisms by processing vast amounts of data generated by physical assets, extracting valuable insights and making predictions that drive operational improvements~\cite{rathore2021role}. Additionally, AI-enabled systems can make real-time decisions based on simulation results and sensor data, allowing for immediate adjustments to operations~\cite{rathore2021role}. This autonomy enhances both efficiency and responsiveness, making DTs more effective in managing and optimizing complex systems.

The integration of \MS{} and AI within DTs fosters advanced system capabilities. For example, surrogate modeling uses AI to create simplified models that approximate complex simulations~\cite{Ender2015} and by that, speed up computation at acceptable costs. AI-augmented simulations predict outcomes quickly using partial data~\cite{adhikari2022comprehensive}. AI also aids in uncertainty quantification~\cite{abdar2021review}. Closed-loop learning allows DTs to adapt by continuously integrating real-time data~\cite{moya2022digital}. These concepts collectively enhance the efficiency and effectiveness of DTs in managing complex systems.

The fusion of \MS{} and AI \emph{supercharges} DTs, enabling them to simulate, learn, and adapt in real time. \MS{} provides the foundation for understanding and testing system behavior, while AI adds intelligence through prediction, autonomy, and continuous learning. This synergy transforms DTs into powerful tools for optimizing complex systems efficiently and proactively.
\index{digital twin|)}

\section{\MS{} in Digital Twins}\label{sec:ms}

DTs represent a paradigm shift where the physical and virtual worlds coalesce to enhance both decision-making and asset management. At the core of this transformation lies \MS{}, which forms the fundamental engine for translating real-world dynamics into meaningful digital representations. Through \MS{}, engineers can virtually experiment with assets under a wide range of scenarios, such as predicting failures, optimizing performance, and enabling proactive maintenance -- all while mitigating risks associated with real-world testing~\cite{boschert2016digital}.
This section provides an overview of how \MS{} acts as the bridge between the physical and virtual worlds.

\subsection{Role and purpose of \MS{}}

\MS{} serve as the conceptual and computational bridge between real-world systems and their digital replicas. Faithful models allow the DT to act as a proxy to physical assets and mirror their structure, dynamics, and control logic. Modeling is the semantic glue that transforms real-time data into actionable virtual replicas that evolve with their physical replicas~\cite{gabor2021digital}.
Simulation is the computational enactment of models under varying conditions and by that, it enables the investigation of various behavioral traces of models and what they represent, i.e., physical assets. Such scenarios are of particularly high utility in the lifecycle management of manufacturing systems as they enable what-if scenario analysis and predictive analytics~\cite{negri2017review}. Additional well-established use-cases of simulation in DTs include risk assessment, trade-off analysis, early validation~\cite{madni2019leveraging}.

\subsection{Integration of \MS{} within the digital twin framework}

The integration of \MS{} in DTs is mostly approached the conceptual level. In building engineering, \citet{zhang2025review} emphasize integrating interoperable data models with simulation workflows to improve adaptability. \citet{mittal2025digital} showcase reduced-order models embedded in real-time DTs of power systems.
Additionally, \citet{hosseini2025milk} demonstrate a hybrid approach capitalizing on discrete-event simulation and optimization for supply chain planning. An illustrative example of a hybrid \MS{} approach is presented by Mykoniatis and Harris~\cite{mykoniatis2021digital} in their DT emulator for a modular production system. Their work combines two \MS{} paradigms: discrete event simulation (DES) to model the overall flow of workpieces through the system, and agent-based models (ABM) to capture the detailed states and behaviors of each individual manufacturing station. The resulting hybrid model is used for virtual commissioning, allowing for the testing and validation of the system's Programmable Logic Controller (PLC) in a simulated environment before it is deployed on the physical hardware. This work exemplifies how different \MS{} formalisms can be combined to create a high-fidelity, functional virtual replica for system validation and optimization.

\index{digital twin!ISO 23247|(}
However, the importance of \MS{} in DTs necessitates proper architectural integration guidelines.
Most often, DTs follow either layered architectural design or a component-based approach, with occasional cases of microservice-based integration~\cite{kosse2025semantic}.
Recognizing the need for standardized integration, the International Organization for Standardization (ISO) released the Digital twin framework for manufacturing -- ISO 23247-1:2021 standard~\cite{iso23247:2021}, which defines a reference framework for DTs in manufacturing. In the entity-based architectural layout of the standard, the role of the simulation functional entity is to ``predicts the behavior of Observable Manufacturing Element (OMEs)''~\cite{shao2024manufacturing}, and it closely integrates with application and service functional entities, such as the ones for analytic services and for reporting. ISO 23247-1:2021 emphasizes that simulation models must be modular, reusable, and dynamically updatable to support life cycle operations along the replicated systems. Although originally developed for the manufacturing domain, the reference architecture of the ISO 23247 standard is an appealing feature for DT developers in other domains, too. Thus, adaptations of ISO 23247 to other domains have already begun, e.g., in space systems~\cite{shtofenmakher2024adaptation}, AI simulation~\cite{liu2025ai}, and automotive systems~\cite{ramdhan2025engineering}. These works underscore the utility of the ISO 23247 reference architecture outside manufacturing and make a clear case for clear architectural design over conceptual integration.
Complementing ISO 23247, the ISO/IEC 30173:2023 -- Digital twin standard~\cite{iso30173:2023} provides a broader systems-level overview, specifying that \MS{} services in DTs should be interoperable across domains and capable of supporting real-time feedback loops, predictive behavior, and semantic context-awareness.

An example implementation of ISO 23247 is presented by \citet{cabral2024digital} for a robotic manufacturing cell. Their DT explicitly maps the physical components of the cell (robot, welding torch, positioning table) to the \emph{Observable Manufacturing Element} (OME).
This structured approach enables the creation of a reliable DT that provides near real-time 3D simulations and process data visualization. This work serves as a good example of how applying a standardized framework facilitates the systematic development of DTs in a modern manufacturing context.

Collectively, these academic contributions and standards affirm that modeling and simulation are not peripheral aspects of DTs but rather, architecturally embedded services that operationalize DTs by transforming data into predictive, interactive, and semantically rich system behavior.\index{digital twin!ISO 23247|)}

\subsection{Modeling and simulation techniques}

After fitting \MS{} into the DT, one has to decide which modeling and simulation formalism to use to capture the essential properties of the physical asset. An apt advice is given by multi-paradigm modeling (MPM), which suggests to \textit{model everything explicitly, at the most appropriate level(s) of abstraction, using the most appropriate formalism(s)}~\cite{vangheluwe2002introduction}. Indeed, choosing the right modeling formalism and simulation technique is the key to a faithful and high-fidelity DT.
In the following, we discuss some of the commonly encountered modeling and simulation methods in DTs.

\index{physics-based modeling|(}
\subsubsection{Physics-based modeling}

Physics-based modeling builds on fundamental principles of physics to describe a system. These models are governed by laws of physics, e.g., conservation of mass, momentum, and energy, and are typically implemented in differential equations, finite element models (FEM)~\cite{dhatt2012finite} or computational fluid dynamics~\cite{anderson1995computational} yield high-fidelity representations that are crucial when spatial variations and transient effects play key roles. Physics-based models are especially valuable when empirical data is limited or when high-fidelity simulations are needed to capture the physical properties of a system.

For example, a physics-based model can be used in the DT of a marine towed cable-body system to simulate cable dynamics via FEM, allowing for precise prediction of structural responses in dynamic marine environments~\cite{westin2025framework}. Such a DT enhances situational awareness and predictive maintenance capabilities.
\index{physics-based modeling|)}

\index{Discrete Event Simulation (DES)|(}
\subsubsection{Discrete event simulation}

A discrete event simulation (DES) model~\cite{varga2001discrete} represents system behavior as a series of discrete events. By capturing the timing and sequence of events, e.g., task completions, resource transitions, or system breakdowns, this approach is effective in modeling process flows and operational scheduling. Its utility lies in its simplicity to represent a system's state over time while still allowing for insights into temporal properties~\cite{wainer2017discrete}.

For example, DES can be used in DTs for the commissioning of automotive production lines~\cite{morabito2021discrete}. Such a DT would process a real-time data stream as input to the simulation, and allow engineers to run instantaneous what-if scenarios to quantify properties of interest.
Another example application of DES is the simulation of building-related IoT infrastructures for what-if scenario modeling of water supply infrastructures~\cite{simultech25}.
\index{Discrete Event Simulation (DES)|)}

\index{Discrete Event System Specification (DEVS)|(}
\subsubsection{Discrete Event System Specification}

Discrete Event System Specification (DEVS)~\cite{zeigler2018theory} is a hierarchically compositional formalism to describe and analyze reactive systems. It allows for constructing system specifications of arbitrary complexity through the sound composition of more primitive DEVS models. DEVS is closed under composition, i.e., composing two DEVS models results in a new DEVS model.
DEVS has been widely regarded as the common denominator among modeling formalisms, some dubbing it as the assembly language of simulation~\cite{vangheluwe2000devs}. Indeed, although a discrete formalism by nature, DEVS supports modeling and simulating continuous state systems and hybrid continuous-discrete state systems.
Compared to DES, DEVS only considers events of interest, i.e., there is no time function with equidistant values (``ticks''); instead, events happen upon triggers internal or external to the DEVS model.
Due to its versatility, DEVS is especially suitable for building simulators for DTs~\cite{david2024automated}.

For example, DEVS can be used to model operation modes of complex equipment (e.g., HVAC, lights, temperature) in production facilities~\cite{david2023digital}. Such a DT would process real-time data from various sensors across the plant and channel them to the DEVS simulator as input. The DEVS model, in turn, allows for simulating future states of the facility as the function of various settings of the equipment.
\index{Discrete Event System Specification (DEVS)|)}

\index{System Dynamics (SD)|(}
\subsubsection{System dynamics}

System dynamics (SD)~\cite{bala2017system} models systems as continuous, feedback-driven structures composed of stocks, flows, and causal loops. Rooted in systems theory, SD is particularly effective for modeling high-level system behavior with time delays and nonlinearity. These models facilitate a macro-level understanding of evolving dynamics, making them well-suited for strategic planning in domains, e.g., energy systems, logistics, and healthcare, and for multisystemic, large-scale problems, e.g., systems sustainability~\cite{manellanga2024participatory}.
In DTs, SD is often used for scenario exploration and long-term policy evaluation.

For example, the DT of an enterprise could use SD for modeling the effects of strategic decisions and by that, improve decision-making~\cite{yan2021integrated}.
\index{System Dynamics (SD)|)}

\index{Agent-based model (ABM)|(}
\subsubsection{Agent-based modeling}

Agent-based models (ABM) decompose the system into autonomously acting agents---such as humans, machines, or software components---within a shared environment~\cite{macal2009agent}. Each agent follows a set of behavioral rules and may possess learning or adaptation capabilities. ABM is particularly suitable to analyze emergent system behavior arising from agent-level interactions.
The combination of the simplicity of modeling and the ability to investigate particularly complex emerging phenomena render ABM an apt choice for complex real-life problems.

For example, a DT of a healthcare facility could use ABM to model patients that go through emergency processes and medical actors who provide care for patients~\cite{croatti2020integration}. Such a DT can help predict queuing times, resource utilization, and resource bottlenecks. In logistics, ABM can be used in warehouse management DTs to model the influx and efflux of goods~\cite{maka2011agent}, useful, e.g., in predicting peaks and limits in storage capacity and optimizing warehouse operations accordingly.
\index{Agent-based model (ABM)|)}

\subsubsection{Hybrid simulation}

Hybrid simulation combines multiple modeling paradigms---such as DEVS, SD, ABM, and physics-based models within a unified framework~\cite{mustafee2020hybrid}. This integrative approach capitalizes on the strengths of each technique while compensating for their individual limitations, thus enabling comprehensive modeling of complex, multi-domain systems. As such, it is a practical embodiment of the multi-paradigm modeling principles~\cite{vangheluwe2002introduction} mentioned earlier in this section.

In DTs, hybrid simulation is essential for modeling complex systems that span multiple abstraction levels or domains. For example, \citet{adams2022hybrid} propose a framework that integrates physics-based and data-driven components, creating hybrid DTs that combine deterministic system models with machine learning for enhanced adaptability and scalability. \citet{nguyen2022interfaces} provide a structured methodology for interfacing SD and ABM, illustrating how hybrid models facilitate richer insights and more robust simulations of adaptive systems. \citet{langlotz2022concept} implement a hybrid-modeled DT for energy management in manufacturing systems, combining physical process modeling with data-driven optimization to coordinate battery usage across factory operations.

Hybrid modeling also aligns with evolving standardization efforts. For example, the ISO/IEC 30173:2023 -- Digital twin standard~\cite{iso30173:2023} recommends that DT simulation services support modular integration of heterogeneous modeling paradigms. This ensures that DTs can provide robust, predictive, and semantically rich representations across varied application domains.

\subsection{Challenges in deploying \MS{} in digital twins}

While DTs offer promising capabilities across engineering, manufacturing, and infrastructure systems, several challenges hinder the effective integration of \MS{} into DTs and their joint deployment. Below, we review some of these challenges.

\begin{description}
    \item[\textbf{Real-time data integration}] Unlike traditional simulation tools that operate on static input datasets, DTs must ingest, process, and synchronize real-time sensor data from the physical asset. Achieving seamless data fusion, low-latency updates, and semantic consistency between the physical and virtual entities remains a persistent bottleneck~\cite{zhou2024data}. Failure to address this can result in degraded predictive power and unreliable decision support.

    \item [\textbf{Testability and calibration}] Testing and calibrating models is not trivial in DTs, especially in cases when safety-critical deployments or observability challenges hinder the process~\cite{david2023digital}. Adaptive techniques, e.g., via reinforcement learning~\cite{david2024automated} offer upside, but systematic methods and protocols for model calibration remain sought-after.

    \item[\textbf{Nonstationarity and model drift}] As physical systems evolve due to wear, maintenance, or operational changes, the simulation model embedded within the DT must adapt dynamically. Static models quickly become obsolete, necessitating online learning or recalibration mechanisms to cope with concept drift~\cite{zhou2024data}. This challenge is particularly acute in data-driven DTs, where the underlying statistical assumptions may no longer hold over time. Quantifying the simulation vs.~reality gap thus is of prime relevance to provide tooling to assess model drift~\cite{zech2025bs}.

    \item[\textbf{Hybrid model complexity}] DTs frequently require the integration of multiple simulation paradigms, e.g., DEVS, SD, ABM, and physics-based models. Orchestrating these heterogeneous approaches within a unified simulation loop is computationally intensive and demands sophisticated middleware for time coordination, state synchronization, and result fusion~\cite{nguyen2022interfaces,langlotz2022concept}.

    \item[\textbf{Fault injection and functional safety}] In safety-critical applications, it is essential to assess how the system behaves under various fault conditions such as sensor failures, actuator anomalies, or cyber intrusions. Modeling such faults and injecting them into simulation environments adds further complexity to both the modeling effort and the validation protocols~\cite{tosoni2022challenges}. These capabilities are crucial for DTs deployed in cyber-physical production systems and industrial automation.

    \item[\textbf{Interoperability and standardization gaps}] The lack of agreed-upon standards for simulation formats, ontologies, and communication protocols impedes model reuse, composability, and interoperability of DTs~\cite{david2024interoperability}. This leads to vendor lock-in, duplicated effort, and poor scalability across domains. Standards such as ISO 30173 aim to address this gap but remain under-utilized and sporadically adopted in practice~\cite{eslahi2024sustainable}.

    \item[\textbf{Data quality and fragmentation}] Particularly in domains such as construction and urban infrastructure, poor data governance practices, inconsistent naming conventions, and fragmented data sources hinder the development of reliable simulation models. Establishing data validation pipelines, metadata standards, and provenance tracking is essential for trustworthy DT simulations \cite{eslahi2024sustainable}.

    \item[\textbf{Computational load and real-time performance}] High-fidelity models demand significant computational resources, especially when simulating complex mechanical, electrical, or thermal interactions. Ensuring real-time or near-real-time performance under these constraints requires novel strategies such as surrogate modeling, GPU acceleration, or multi-resolution simulation hierarchies.

    \item[\textbf{Lifecycle coverage and model reusability}] Many current DT implementations are restricted to isolated phases of the asset life cycle such as design or operation. Full life cycle modeling (design, build, operate, maintain, retire) remains an open challenge, often due to siloed data, lack of model reusability, or unclear ownership of model updates across organizational boundaries~\cite{zech2024requirements,zech2025agile,zech2025empirical}.

    \item[\textbf{}] 
\end{description}

These challenges emphasize that while the conceptual promise of DTs is strong, their realization in real-world systems must overcome significant theoretical and engineering hurdles. Addressing these issues requires interdisciplinary collaboration, new computational frameworks, and more mature software ecosystems.

\section{AI in Digital Twins}\label{sec:ai}

AI comprises a broad class of computational methods and systems aimed at emulating or augmenting human cognitive abilities such as reasoning, learning, perception, and autonomous decision-making~\cite{jiang2022quo}. Over the past decade, AI has progressed from rule-based expert systems to encompass advanced data-driven paradigms, including machine learning, deep learning, and reinforcement learning, all enabled by increased computational power and the availability of large-scale data~\cite{abdar2021review,bariah2024interplay}.

Contemporary AI algorithms can autonomously extract patterns from complex, high-dimensional structured and unstructured datasets, underpinning applications such as predictive analytics, autonomous control, speech recognition, and natural language processing~\cite{garg2021overview}. The integration of AI yields significant advantages, notably where analytical models are difficult to construct or where systems must operate under uncertainty and evolving dynamics~\cite{rathore2021role,bariah2024interplay}. Therefore, AI operates as a key analytical driver, offering predictive and prescriptive insights that support automation, optimization, and resilience in complex systems.

\subsection{The Roles of AI in Digital Twins}

Fully leveraging the potential of DTs---particularly in terms of automation, intelligence, and adaptability---requires the integration of AI techniques. Not only does AI augment the predictive and analytical capabilities of DTs, but also enables a shift toward more autonomous, self-optimizing, and context-aware systems~\cite{rathore2021role}.

\begin{figure}
    \centering
    \includegraphics[width=0.9\linewidth]{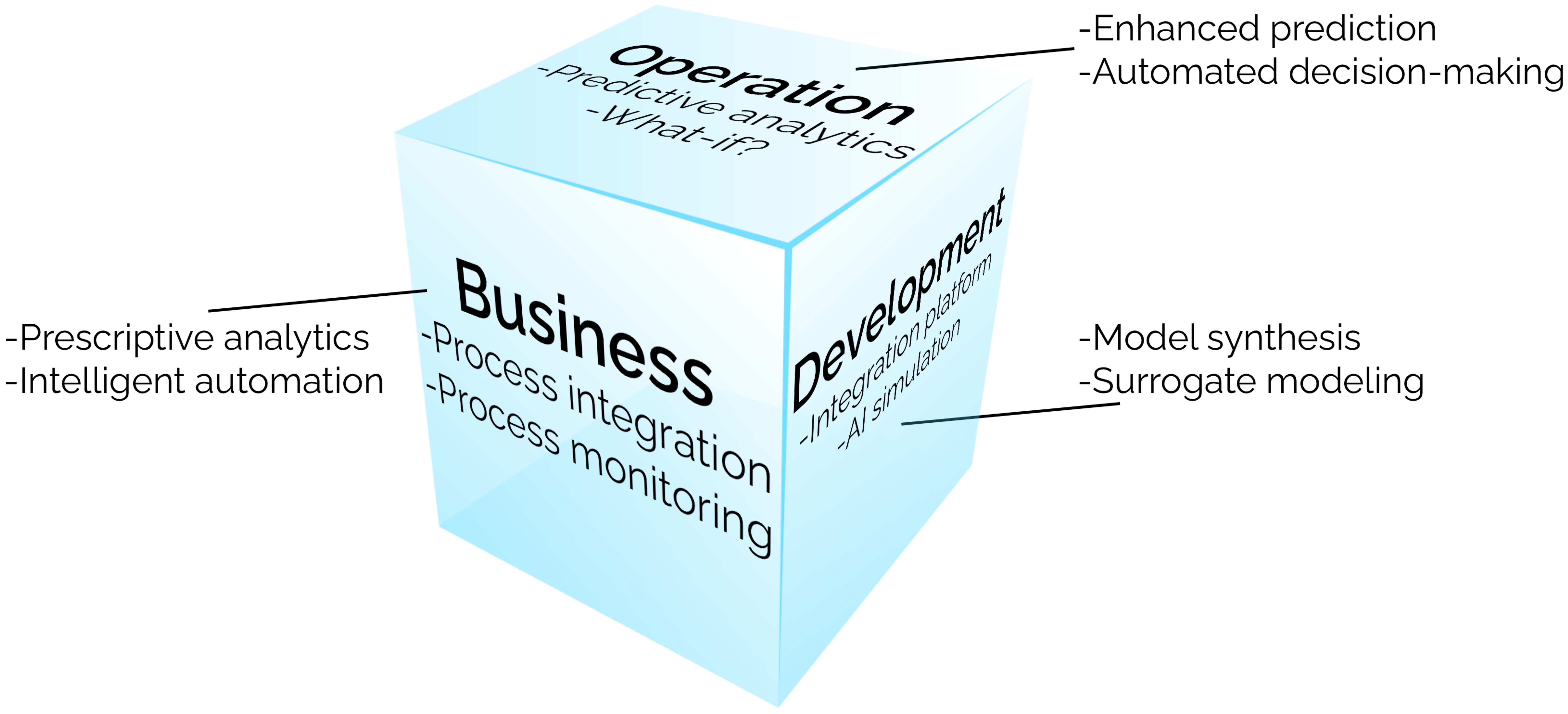}
    \caption{Typical roles of AI in the different views of DTs}
    \label{fig:cube-ai}
\end{figure}

To understand the role of AI in DTs, we relate AI to the operational, development, and business aspects of DTs. Some of the key examples are shown in \figref{fig:cube-ai}. 

\subsubsection{Operation}
\begin{description}
    \item[\textbf{Enhancing prediction}] Machine learning models analyze high-dimensional data streams from the physical asset to identify subtle patterns that may precede a fault. This data-driven approach enables a shift from reactive to predictive maintenance, which reduces operational downtime and extends the asset's lifespan~\cite{chen2023advance}.
    \item[\textbf{Automated decision-making}] The DT provides a risk-free simulation environment for training reinforcement learning (RL) agents. These agents learn optimal control policies through trial and error, which can then be deployed to the physical asset for real-time process optimization and autonomous adaptation~\cite{santos2022use}.
\end{description}
    
\subsubsection{Development}
\begin{description}
    \item[\textbf{Model synthesis}] AI techniques can automate the creation and calibration of simulation models by learning system dynamics directly from sensor data. This continuous, automated calibration ensures that the DT remains a faithful representation of the physical asset as it evolves over time~\cite{akhavan2024generative}.
    \item[\textbf{Surrogate modeling}] Computationally expensive, high-fidelity simulations can be replaced by AI-based surrogate models, such as deep neural networks. These surrogates provide predictions orders of magnitude faster, enabling the use of complex models in real-time control loops and decision support systems~\cite{vsturek2025surrogate}.
\end{description}
    
\subsubsection{Business}
\begin{description}
    \item[\textbf{Prescriptive analytics}] By integrating predictive insights with optimization algorithms, AI enables the DT to evaluate numerous future scenarios and recommend the optimal course of action required to achieve specific business objectives~\cite{gottimukkala2025prescriptive}.
    \item[\textbf{Intelligent automation}] Combining process mining with intelligent agents allows the DT to link physical operations with their corresponding business workflows. This integrated view enables the automation of complex decisions and the extraction of actionable knowledge for continuous process improvement~\cite{jazdi2021realization}.
\end{description}

\subsection{The Role of Digital Twins in AI}

While AI serves as a powerful enabler of DTs, the relationship is bidirectional: DTs also help advance AI. DTs often act as training environments as well as a deployment platform for AI systems.

\subsubsection{DTs as safe and cost-efficient training environments}
A key challenge in AI is the need for high volumes of high-quality data (e.g., in deep learning) or safe environments for exploration (e.g., in reinforcement learning). The high-fidelity simulators of DTs allow them to act as data generation end-points or training environments that complement or completely substitute scarce, privacy-sensitive, or costly real-world datasets. For example, a physics-based simulation in a DT can produce data for machine learning, including rare or extreme scenarios (e.g., safety-critical failures) that would be impractical to observe in reality. Another pertinent example is training reinforcement learning agents in sandboxes provided by DTs, where agents can interact with realistic system dynamics, learn policies through trial-and-error, and transfer their knowledge to the physical system efficiently, thanks to the strongly coupled physical twin. \citet{liu2026developing} provide details of such cases and solution patterns.

\subsubsection{DTs for validation and testing}
The integration of AI into critical domains such as aerospace, energy, healthcare, or manufacturing demands rigorous testing before deployment. DTs enable the verification and validation (V\&V) of AI models, e.g., by stress-testing them against simulated conditions to ensure reliability, robustness, and compliance with safety standards.

\subsubsection{Accelerating AI development cycles}
DTs help shorten AI development lifecycles in at least two ways. First, DTs allow for scalable realistic experimentation, enabling large-scale experimentation and hyperparameter tuning without disrupting real-world operations. Second, through the real-time connection to the physical system, DTs allow for the continuous update and maintenance of AI models in near real time. DTs with such continuous learning capabilities are often referred as cognitive DTs.
    
\subsubsection{Bridging the sim-to-real gap}
A key challenge in AI is the simulation-to-reality (sim-to-real) gap, i.e., when models trained in synthetic environments underperform in real-world conditions. DTs help mitigate this issue by seamlessly deploying, testing, and fine-tuning the trained AI model on the physical asset. By that, DTs reduce discrepancies between virtual and real domains, ensuring the validity of AI models when transferred to the operational environment.
    
\subsubsection{DTs as integration and deployment platforms for AI systems}
Beyond serving as environments for training and testing, DTs also function as integration and deployment platforms for AI systems. In this role, the DT acts as a technical framework to integrate human-engineered and AI-inferred models, typically by packaging models and simulators behind standardized interfaces, such as the Functional Mockup Interface.\footnote{\url{https://fmi-standard.org/}} Conceptually, the DT in this role also acts as an intermediary between the physical asset and AI-driven decision-making, enabling safer, transparent, and efficient integration of AI-enabled intelligent functions into real-world operations.

\subsection{The role of data}

The successful utilization of DTs and AI fundamentally relies on the availability, quality, and governance of data.
Both DTs and AI are inherently data-driven: DTs require a steady influx of high-fidelity real-time and historical data to maintain accurate synchronization with their physical counterparts, while AI techniques depend on large, diverse datasets to support tasks such as learning, prediction, and automation. Inadequate data management can undermine the efficacy of both DTs and AI, leading to unreliable models, poor decision support, and diminished operational value~\cite{bertossi2020data}.

To realize the benefits of DTs, especially when complemented by AI and advanced analytics, focus on data aspects is essential. This includes standardized data acquisition, semantic interoperability, data integrity, provenance, and life cycle management. Standards such as ISO 8000 (``Data quality'') emphasize these aspects alongside structured approaches to data representation, exchange, and quality assurance as prerequisites for interoperability and trustworthiness in organizational settings~\cite{iso8000}. Key data management requirements for successful DT and AI implementations include the following.

\begin{description}
    \item[\textbf{Data quality and integrity}] Ensuring the accuracy, consistency, and completeness of streaming and stored data that underpins modeling, analytics, and feedback.
    \item[\textbf{Semantic interoperability}] Adopting shared vocabularies, ontologies, and compliant data formats to enable seamless data exchange between systems and domains.
    \item[\textbf{Provenance and traceability}] Maintaining transparent records of data origin, transformations, and lineage to support audits, validation, and regulatory compliance.
    \item[\textbf{Scalability and accessibility}] Architecting data infrastructures that support efficient handling of large volumes of heterogeneous data and provide secure yet flexible access for analytics and decision-making.
\end{description}

Adhering to such data-centric best practices is essential not only for robust DT deployment, but also for exploiting the full value of AI-enabled automation and optimization in digitalized environments.

\section{By Way of Conclusion}\label{sec:outlooks}

In this chapter, we explored how DTs, \MS{}, and AI align in the design and operation of complex systems. We touched upon the typical roles of DTs, the frequently used \MS{} formalisms and AI techniques, and integration into DTs.
We also touched upon standards that frame and drive DT development and adoption in industry.

We established that DTs play crucial technical, operational, and business roles that make the case for the convergence of \MS{} and AI. As well, DT make use of \MS{} and AI, strengthening the synergy among these techniques.

The techniques discussed in this chapter shape the next generation of modern intelligent systems, drawing significant attention from researchers and adopters alike. To motivate the reader to continue exploring these topics, in the remainder of this chapter, we review the challenges and future research and development directions for \MS{}-driven, AI-enabled DTs.

\subsection{Challenges and recommended future directions}

To leverage the convergence of \MS{}, AI in the context of DTs, there are some challenges to be tackled. The following is a non-exhaustive sample of the long list of such challenges and recommended future directions.

\subsubsection{Seamless integration of \MS{} and AI}

Seamless integration will allow for exploiting the synergies between \MS{} and AI to a greater extent and implement more complex systems. A pertinent example are self-healing and self-calibrating systems enabled by automated model maintenance using reinforcement learning in response to changing physical conditions. To achieve this, unified workflows and architectures are needed that allow human engineering and data-driven AI methods to complement and enhance each other within a single framework. Furthermore, ensuring seamless lifecycle management of models is required---including their calibration, synchronization, continuous validation, and evolution. DTs offer an excellent methodological and technological starting point for such integration efforts.

\begin{recommendation}{Recommendation}%
Develop open, modular DT platforms that natively support hybrid workflows, multi-paradigm modeling lifecycles, and AI model validation.
\end{recommendation}
    
\subsubsection{Model fidelity and computational demands} In simulation, a lot hinges on the fidelity of the simulation model. On the one hand, vague and simplified models will not allow for detailed insights. On the other hand, simplified models allow for faster evaluation and demand less computing resources. Achieving the right balance between high-fidelity simulation and scalable, real-time performance, therefore, is an important challenge. AI-driven surrogate models are prime candidates to tackle such challenges. Adaptively switching between different modeling paradigms that are better-suited for specific problems is a possible direction, too.

\begin{recommendation}{Recommendation}%
    Advance scalable surrogate modeling and automated model reduction and optimization techniques, enabling real-time, high-fidelity simulations that integrate transparently with AI components.
\end{recommendation}
    
\subsubsection{Data-related challenges} Save for online flavors of machine learning (i.e., learning from data that becomes available sequentially), the success of AI methods depends on the volume and quality of training data. Thus, ensuring the availability, quality, and semantic interoperability of data is necessary in AI-enabled \MS{}. A particular challenge within this realm is the systematic handling and imputation of incomplete or fragmented datasets, preferably through DTs~\cite{liu2026developing}. Another key challenge in this realm is establishing sound data governance methods to support continuous learning, traceability, and trustworthy decision-making in dynamic environments.

\newpage

\begin{recommendation}{Recommendation}%
    Implement rigorous data governance frameworks which are anchored in standards like ISO 8000 to ensure data quality, semantic interoperability, governance, and provenance for all DT and AI operations.
\end{recommendation}
    
\subsubsection{Explainability and trust} One of the impediments to the widespread adoption of modern AI is the lack of transparency of AI components. This, in turn, results in the lack of explainability of AI models' predictions, and ultimately, in the lack of trust by human experts and stakeholders. This is especially problematic when AI-enabled \MS{} and DTs are used in critical systems. Until this challenge is tackled, the convergence of \MS{} and AI will be hindered and limited.

\begin{recommendation}{Recommendation}%
    Design validation and verification methodologies that link explainable AI outputs with simulation results, making hybrid decision processes explainable, traceable, and eventually, trustworthy.
\end{recommendation}
    
\subsubsection{Interdisciplinary collaboration and standardization} Since \MS{} and AI are typically developed and employed at the intersection of diverse communities, interdisciplinary collaborations are conductive to the progress of both \MS{} and AI, as well as their convergence, by yielding unique use cases, problems, and solution patterns. To ease collaborations and cross-disciplinary alignment, adopting robust standards for data, interfaces, and model interoperability seems to be paramount. This will support modular and reusable DT solutions as well.

\begin{recommendation}{Recommendation}%
    Foster communities of practice across \MS{}, AI, and DT engineering to co-develop tools, workflows, and open standards for robust, modular, and reusable DT solutions.
\end{recommendation}

\subsection{Looking ahead}

It is widely expected that AI will profoundly reshape the way we develop, use, and maintain models and simulators. Digital twins close the gap between the digital and physical space, and provide an apt technological integration platform for AI and \MS{}, positioning them as key enablers of future software-intensive systems.

By systematically addressing the challenges outlined above, the integration of \MS{} and AI within DTs will advance both the reliability and autonomy of future systems. Achieving this will demand focused technical contributions, as well as close collaboration across disciplines, alongside the continued development of standards and open platforms.

\section*{Reflection and Exploration}

\begin{questype}{Digital twins}%
    \begin{enumerate}
        \item What is the difference between modeling and a DT? What is the difference between simulation and a DT?
        \item What are the key components of DTs?
        \item What is the key difference between a digital model, a digital shadow, and a digital twin?
    \end{enumerate}
\end{questype}

\begin{questype}{Modeling and AI}%
Think of a recent project of yours or a domain you are familiar with (e.g., automotive, manufacturing, healthcare).
    \begin{enumerate}
        \item Choose a \MS{} formalism from \secref{sec:ms}, preferably one you are familiar with, and explain how that formalism could be used in a DT in your project or domain.
        \item Identify a case that would be hard to model manually, making the case for machine learning and AI models.
        \item Choose an AI technique from \secref{sec:ai}, preferably one you are familiar with, and explain how that technique could solve the problem you identified in the previous point.
    \end{enumerate}
\end{questype}

\begin{questype}{Challenges}%
Elaborate on two-three challenges in the convergence of \MS{} and AI in the context of DTs.
\end{questype}

\begin{exploration}{Further exploration}%
For further exploration, the reader is referred to the following studies that focus on the synergy of \MS{}, AI, and DTs.
    \begin{enumerate}
        \item ``Surrogate Modeling: Review and Opportunities for Expert Knowledge Integration''~\cite{vsturek2025surrogate}
        \item ``Automated Inference of Simulators in Digital Twins''~\cite{david2024automated}
        \item ``AI Simulation by Digital Twins: Systematic Survey, Reference Framework, and Mapping to a Standardized Architecture''~\cite{liu2025ai}
    \end{enumerate}
\end{exploration}

\bibliographystyle{plainnat}
\bibliography{references}

\end{document}

%% file: preamble.tex
\usepackage{type1cm}
\usepackage{makeidx}
\usepackage{graphicx}
\usepackage{multicol}
\usepackage[bottom]{footmisc}
\usepackage{newtxtext}
\usepackage[varvw]{newtxmath}

\usepackage[utf8]{inputenc}
\usepackage{booktabs}
\usepackage[numbers,sort&compress,comma,square,sectionbib]{natbib}
\usepackage{chapterbib}
\usepackage{doi}
\usepackage[dvipsnames]{xcolor}
\usepackage{subcaption}
\usepackage[thinc]{esdiff}
\usepackage{float}

\newcommand{\ms}{modeling \& simulation}
\newcommand{\MS}{M\&S}
\newcommand{\coo}{\ensuremath{\mathrm{CO_2}}}

\newcommand{\secref}[1]{Section~\ref{#1}}

\newcommand{\figref}[1]{Figure~\ref{#1}}

\definecolor{recommendation}{gray}{0.8}
\newenvironment{recommendation}[1]{\ignorespaces\def\stmtopen##1{##1}%
\formtmp{recommendation}{#1}}{\par\noindent\textcolor{recommendation}{\rule{\columnwidth}{1pt}}\vskip2pt\par\addvspace{\baselineskip}}%

\definecolor{exploration}{gray}{0.8}
\newenvironment{exploration}[1]{\ignorespaces\def\stmtopen##1{##1}%
\formtmp{exploration}{\ \circledmark{\textcolor{black}{$~\blacktriangleright$}}{\kern6pt}#1}}{\par\noindent\textcolor{exploration}{\rule{\columnwidth}{1pt}}\vskip2pt\par\addvspace{\baselineskip}}%